\documentclass[10pt,twocolumn,letterpaper]{article}
\usepackage{wacv}
\usepackage{amsmath,amsfonts,bm,booktabs,multirow,array,tabularx}
\usepackage{mathtools}
\usepackage{algorithm,algorithmic}
\usepackage{xcolor}
\usepackage{microtype} 

\definecolor{wacvblue}{rgb}{0.21,0.49,0.74}
\usepackage[pagebackref,breaklinks,colorlinks,allcolors=wacvblue]{hyperref}
\def\wacvPaperID{2651}
\def\confName{WACV}
\def\confYear{2027}

\title{RigidDriveGS: Object-Centric Vehicle Reconstruction from Fixed Multi-Camera Drive-Through Captures\vspace{-0.5 em}}

\author{
Nitin Kulkarni\textsuperscript{1,2},
Akhil Devarashetti\textsuperscript{1},
Charlie Cluss\textsuperscript{1},
Livio Forte\textsuperscript{1},\\
Philip Schneider\textsuperscript{1},
Chunming Qiao\textsuperscript{2}\thanks{%
Chunming Qiao is supported in part by the National Science Foundation
under Grant Nos.\ CNS-2413876 and CNS-2120369.
The authors thank ACV Auctions for providing computing resources,
vehicle data, and the camera rig.},
Alina Vereshchaka\textsuperscript{2}\\[0.5em]
\textsuperscript{1}ACV Auctions, Buffalo, NY, USA\\
\textsuperscript{2}University at Buffalo, Buffalo, NY, USA\\[0.3em]
{\normalfont\small
\{adevarashetti, ccluss, lforte, pschneider\}@acvauctions.com}\\
{\normalfont\small
\{nitinvis, qiao, avereshc\}@buffalo.edu}\vspace{-0.5 em}
}

\begin{document}
\maketitle

\begingroup
\renewcommand{\thefootnote}{}
\footnotetext{%
\scriptsize
\textcopyright{} 2027 IEEE. Personal use of this material is permitted.
Permission from IEEE must be obtained for all other uses, in any
current or future media, including reprinting/republishing this
material for advertising or promotional purposes, creating new
collective works, for resale or redistribution to servers or lists,
or reuse of any copyrighted component of this work in other works.
}
\endgroup

\setlength{\abovedisplayskip}{3pt}
\setlength{\belowdisplayskip}{3pt}
\setlength{\abovedisplayshortskip}{2pt}
\setlength{\belowdisplayshortskip}{2pt}
\setlength{\jot}{1pt}

\begin{abstract}
High-fidelity 3D reconstruction of vehicle exteriors improves buyer confidence in online automotive marketplaces, but fixed multi-camera drive-through capture departs from standard Structure-from-Motion (SfM): static cameras observe a moving vehicle amid clutter, rotating wheels, fisheye distortion, and specular reflections. We present \mbox{RigidDriveGS}, an object-centric pipeline combining motion-gated tracking, bifurcated masks that exclude wheels for SfM while optionally retaining them for rendering, mask-guided dense correspondences on raw distorted images, rig-constrained global SfM, and clean object-centric neural rendering with trimap-aware opacity or density supervision and native-alpha white compositing. We evaluate 25 vehicles from 10 dealership installations and six phone-recorded vehicles, each captured from 14 independent handheld positions. Under a staggered per-camera temporal-block holdout, 3DGUT+MCMC achieves $30.42$ dB PSNR, $0.90$ SSIM, and $0.20$ LPIPS, improving PSNR by $7.78$ dB over undistorted 3D-GS; independently recorded reference LiDAR scans yield full/body Chamfer-L1 of $0.026/0.013$ m and F1@5 cm of $0.703/0.885$. These results support clutter-free, object-centric novel-view synthesis without controlled studio infrastructure.
\end{abstract}

\section{Introduction}
\label{sec:introduction}

Online vehicle marketplaces and wholesale auctions rely on sparse web-listed imagery with limited viewpoint coverage. Object-centric 3D reconstruction enables continuous walk-around visualization. We address reconstruction and novel-view synthesis, rather than defect detection or metric damage assessment.

Although advances such as 3D Gaussian Splatting (3D-GS) enable interactive neural rendering in robotics and vision systems, including SLAM and calibration pipelines~\cite{sun2024mm3dgs,herau20243dgs}, vehicle reconstruction often assumes controlled photo booths or turntables. More scalable fixed drive-through rigs invert conventional SfM: static cameras observe a moving vehicle, making the cluttered dealership background an attractive but incorrect reconstruction target~\cite{schonberger2016structure}. Short captures, fisheye distortion, specular paint and glass, independently moving objects, and rotating wheels further challenge correspondence estimation and rendering.

We propose \textbf{RigidDriveGS}, an object-centric pipeline that preserves native distorted imagery while separating rigid geometric support from the complete vehicle silhouette. Our contributions are: (i) motion-gated mask decomposition into rigid SfM support, optional complete appearance support, and independent privacy exclusion; (ii) mask-guided dense correspondence sampling on full, raw distorted images; (iii) object-centric global SfM with CAD-derived rig priors and bounded extrinsic refinement; and (iv) clean object-centric rendering from unmodified RGB using separate supervision for foreground, uncertain boundaries, and confident exterior regions through native alpha or transmittance.

We evaluate 25 vehicles captured by synchronized \mbox{14-camera} rigs at 10 dealerships and six vehicles recorded from 14 independent handheld positions. A staggered temporal-block holdout evaluates appearance, while independently recorded LiDAR scans provide a surface reference for the handheld set. Fixed-rig optimization uses CAD-derived relative-camera priors, whereas handheld passes are reconstructed without relative-pose priors. Applying the pipeline independently to every recording evaluates robustness across vehicles, capture layouts, and dealership environments.

\newpage\clearpage
\section{Related Work}
\label{sec:related_work}

\subsection{Object-Centric Reconstruction}
Multi-view reconstruction typically assumes a moving camera and static scene or controlled object capture~\cite{schonberger2016structure,jensen2014large}. Fixed drive-through rigs offer scalable vehicle capture~\cite{kulkarni2025rig} but invert this geometry: static cameras observe a moving target amid clutter. Unlike dynamic-scene SLAM methods that remove moving objects to estimate camera motion or a static map~\cite{bescos2018dynaslam}, the moving vehicle is our reconstruction target. Prior fixed-rig undercarriage capture~\cite{kulkarni2025rig} is target-dominated and permits static rig-shell masks. Exterior capture additionally involves time-varying clutter, independently moving distractors, reflective and symmetric surfaces, and rotating wheels. These conditions motivate separating the approximately rigid SfM support from the complete rendering silhouette.

\subsection{Learned Matching under Wide-FOV Distortion}
Handcrafted features such as SIFT and ORB are brittle under large viewpoint changes, fisheye distortion, weak texture, and specular reflections~\cite{lowe2004distinctive,rublee2011orb}. Learned methods including DISK+LightGlue, LoFTR, and RoMA improve robustness through learned correspondence priors and broader visual context~\cite{tyszkiewicz2020disk,lindenberger2023lightglue,sun2021loftr,edstedt2024roma,edstedt2025roma}. In drive-through capture, pre-undistortion can discard peripheral detail, while textured stationary backgrounds can dominate matching. We therefore retain the complete calibrated image for inference and use rigid-body masks for endpoint admission and confidence-aware post-filtering. All matcher baselines share the same images, masks, pair graph, intrinsics, geometric verification, and mapper.

\subsection{Neural Rendering}
Neural Radiance Fields provide high-quality novel views but can be costly for high-throughput workflows~\cite{mildenhall2021nerf}; Instant-NGP provides a faster radiance-field baseline~\cite{muller2022instant}. 3D Gaussian Splatting offers an explicit real-time representation~\cite{kerbl20233d}, but its standard pinhole projection is mismatched to short-stand-off fisheye capture. Fisheye-GS and 3DGUT model non-linear projections directly~\cite{liao2024fisheye,wu20253dgut}, while 2DGS targets surface-aligned geometry~\cite{huang20242dgs}. Reflective appearance also motivates regularization and stochastic densification~\cite{li2025gaussian,kheradmand20243d}. We couple these renderers to object-centric SfM through a common clean-background objective that supervises unmodified foreground RGB, treats uncertain boundaries softly, suppresses confident exterior opacity or density, and composites native predictions over white. The distinction is between renderer architecture and its supervision: retaining raw RGB avoids teaching synthetic white boundaries, while separate foreground, boundary, and exterior terms control support. We compare renderers under frozen object-centric poses rather than changing their reconstruction geometry.

\begin{table}[t]
\centering
\footnotesize
\caption{Calibration and capture geometry for synchronized fixed-rig and independent handheld acquisition.}
\label{tab:regimes}
\begin{tabular}{@{}
    >{\centering\arraybackslash}m{0.22\columnwidth}
    >{\centering\arraybackslash}m{0.31\columnwidth}
    >{\centering\arraybackslash}m{0.37\columnwidth}
    @{}}
\toprule
 & Fixed rig & Handheld transfer \\
\midrule
\textbf{Acquisition}
    & 14 fixed, synchronized cameras
    & 14 independent handheld passes \\
\addlinespace[3pt]
\textbf{Intrinsics}
    & ChArUco; 8-param. fisheye
    & ChArUco; $k_1{+}k_2$, zero tangential \\
\addlinespace[3pt]
\textbf{Extrinsics}
    & CAD priors + bounded refinement
    & Unknown; object-centric SfM \\
\addlinespace[3pt]
\textbf{\shortstack{Reconstruction\\scale}}
    & CAD relative translations
    & Global similarity gauge \\
\addlinespace[3pt]
\textbf{Surface reference}
    & None
    & LiDAR, evaluation only \\
\midrule
\textbf{Shared stages}
    & \multicolumn{2}{
        >{\centering\arraybackslash}m{0.69\columnwidth}@{}
      }{Masks, matching, object-centric SfM, rendering} \\
\bottomrule
\end{tabular}
\end{table}

\section{Methodology}
\label{sec:methodology}
\begin{figure*}[t]
  \centering
  \includegraphics[width=\textwidth]{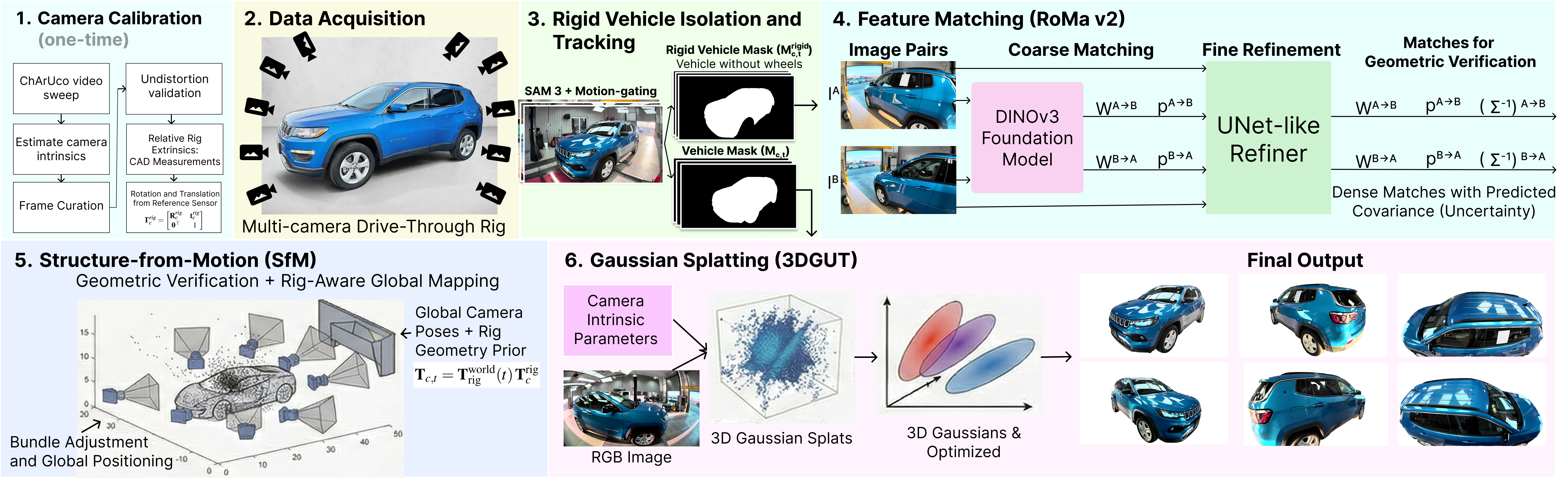}
  \caption{RigidDriveGS model pipeline. Raw frames are tracked and masked, dense matches are admitted on rigid vehicle endpoints, object-centric SfM estimates a vehicle-centric scaffold, and distortion-aware splatting reconstructs appearance. The synchronized fixed-rig branch uses CAD pose priors; the handheld branch replaces this block with rig-free global SfM over 14 independent passes.}
  \label{fig:pipeline}
\end{figure*}

\subsection{Camera Calibration and Capture Geometry}
\label{subsec:methodology_camera_calibration}
Figure~\ref{fig:pipeline} and Table~\ref{tab:regimes} summarize two acquisition branches that share target isolation, matching, SfM verification, and rendering. Both are calibrated offline using a large-format ChArUco board~\cite{an2018charuco}, with camera models appropriate to their optics. The fixed wide-angle cameras use an eight-parameter fisheye model in the raw image domain, whereas the handheld camera uses a central radial model with two radial-distortion coefficients and zero tangential distortion. The calibrated intrinsics remain fixed during matching, bundle adjustment, and rendering, avoiding repeated resampling and maintaining a consistent projection domain throughout the pipeline.

Fixed-rig videos are hardware synchronized. A reference calibration and CAD-derived camera arrangement are transferred across installations. We use column vectors and write $\mathbf T_{a\leftarrow b}$ for the transform that maps coordinates from frame $b$ to frame $a$. For camera $c$, $\bar{\mathbf T}_{c\leftarrow r}\in\mathrm{SE}(3)$ is the nominal CAD transform from the static rig frame $r$ to the camera frame $c$. Small installation tolerances are absorbed by a local perturbation
\begin{equation}
\mathbf T_{c\leftarrow r}
=\exp\!\left(\widehat{\boldsymbol\xi}_c\right)\bar{\mathbf T}_{c\leftarrow r},
\qquad \boldsymbol\xi_c\in\mathbb R^6,
\label{eq:rig_extrinsic_refinement}
\end{equation}
regularized by the CAD-derived prior
\begin{equation}
\mathcal L_{\mathrm{rig}}=\lambda_{\mathrm{rig}}\sum_c
\boldsymbol\xi_c^{\top}\boldsymbol\Sigma_c^{-1}\boldsymbol\xi_c,
\end{equation}
where $\boldsymbol\Sigma_c$ encodes pose confidence. In the handheld branch, every clip is an independent pass: neither phone pose nor vehicle world pose is assumed stable across clips, and inter-clip extrinsics are recovered from verified rigid-vehicle correspondences by rig-free object-centric SfM.

The calibration board contains $53{\times}37$ inner squares, with 22-mm square sides and 16-mm markers. Calibration sweeps vary roll, pitch, yaw, and distance over approximately \mbox{30--120} cm to sample the camera field of view. ArUco IDs disambiguate the board observations. Supplement Sec.~S1 gives calibration and projection details.


\newpage
\subsection{Rigid Vehicle Isolation and Tracking}
\label{subsec:methodology_rigid_vehicle_isolation_and_tracking}
SfM assumes a static scene and can lock onto the high-gradient background instead of the moving vehicle. We therefore introduce a dynamic-scene disambiguation stage that treats the vehicle as the object-centric reconstruction target and suppresses background structure before correspondence estimation.

\subsubsection{Motion-Gated Target Selection and Anchor-Locked Tracking}
For frame $I_{c,t}$, SAM~3~\cite{carion2025sam} produces candidate vehicle instances $\{S^{(i)}_{c,t}\}$, while MOG2~\cite{zivkovic2004improved} produces a binary motion mask $B_{c,t}$. We define the normalized motion energy as
\begin{equation}
E_{c,t}=|\Omega|^{-1}\sum_{\mathbf u\in\Omega}B_{c,t}(\mathbf u),
\label{eq:motion_energy}
\end{equation}
which identifies the active portion of the recording and proposes candidate anchor frames. For a candidate instance $S$, we compute
\begin{equation}
\begin{aligned}
\mathcal J(S)&=\alpha_{\mathrm{mot}}\operatorname{IoU}(S,B)
+\alpha_{\mathrm{tmp}}\operatorname{IoU}(S,M_{t-1})
+\alpha_{\mathrm{roi}}r_{\mathrm{roi}}\\
&\quad+\alpha_{\mathrm{area}}\log(|S|+1)
+\alpha_{\mathrm{sim}}a_{\mathrm{sim}}
+\alpha_{\mathrm{det}}s_{\mathrm{det}},
\end{aligned}
\label{eq:sam_score}
\end{equation}
where the terms measure motion agreement, temporal agreement, optional region support, candidate area, consistency with the running area reference, and detector confidence. Coefficients and tracking thresholds were selected once by grid search on a validation subset using multi-view track consistency and identity-switch frequency, then frozen for all recordings (Supplement Sec.~S2). No reported SfM or rendering result was used for per-recording retuning.

Starting from the selected anchor, tracking proceeds forward and backward under a temporal lock. A candidate is accepted only when it satisfies the frozen overlap and area-consistency criteria. Short detection gaps permit bounded re-acquisition, whereas an extended sequence of misses terminates tracking in that direction. Frames outside the resulting active interval receive blank masks, preventing switches to parked vehicles when the target is absent or only partially visible. Supplement Sec.~S2 reports the model and numerical operating point.

\subsubsection{Rigid-Body and Rendering Masks}
Let $M_{c,t}$ be the tracked full silhouette and $W_{c,t}$ the wheel region. The geometry-valid mask is
\begin{equation}
M^{\mathrm{rigid}}_{c,t}=M_{c,t}\wedge\neg W_{c,t}.
\label{eq:rigid_body_mask}
\end{equation}
Wheel instances are unioned and dilated once with a $7{\times}7$-pixel elliptical kernel (three-pixel radius) before subtraction. $M^{\mathrm{rigid}}$ restricts matching and SfM to the approximately rigid body, avoiding wheel correspondences that violate a single rigid epipolar model. For rendering, we define the complete silhouette as
\begin{equation}
M^{\mathrm{render}}_{c,t}=M_{c,t},
\label{eq:render_mask}
\end{equation}
which retains the wheels. The controlled renderer comparison uses the rigid mask to hold approximately static support constant across methods. A matched best-renderer ablation instead uses the complete appearance mask to quantify the application tradeoff between complete wheel coverage and articulated-region fidelity. Privacy redactions are independent of these masks and are excluded from correspondence, training, and scoring.

\subsection{Feature Matching}
\label{subsec:methodology_feature_matching}

\subsubsection{Image Pair Graph Generation}
Exhaustive matching across all selected frames scales as $\mathcal O(|\mathcal V|^2)$ and includes many pairs with no shared vehicle visibility. We therefore construct a bounded pair graph $\mathcal G=(\mathcal V,\mathcal E)$ that preserves the spatial and temporal connectivity needed for object-centric SfM. Let $\mathbf A\in\{0,1\}^{C\times C}$ encode permitted cross-camera links. The edge set is
\begin{equation}
\begin{aligned}
\mathcal E&=\mathcal E_{\mathrm{spatial}}\cup\mathcal E_{\mathrm{temporal}},\\
\mathcal E_{\mathrm{spatial}}&=\{(I_{k,t},I_{j,s}):\mathbf A_{k,j}=1,\ s\in\mathcal N^{\mathrm{sp}}_{k,j}(t)\},\\
\mathcal E_{\mathrm{temporal}}&=\{(I_{k,t},I_{k,s}):s\in\mathcal N^{\mathrm{tmp}}_{k}(t)\}.
\end{aligned}
\label{eq:feature_pair_graph}
\end{equation}
Here, ``temporal'' denotes continuity within one physical camera stream, whereas ``spatial'' denotes connectivity between viewpoints. For synchronized capture, $\mathcal N^{\mathrm{sp}}$ is centered on hardware-aligned frames and $\mathcal N^{\mathrm{tmp}}$ contains nearby frames from the same camera. For independent handheld passes, cross-clip timestamps are not comparable: $\mathcal N^{\mathrm{tmp}}$ remains within a clip, while $\mathcal N^{\mathrm{sp}}$ aligns permitted viewpoints by normalized visible-vehicle progress. Both neighborhoods have bounded cardinality, giving $|\mathcal E|=\mathcal O(|\mathcal V|(\kappa_{\mathrm{tmp}}+d\kappa_{\mathrm{sp}}))$ for maximum camera-graph degree $d$. The supplement specifies the fixed graph neighborhoods and phase policy.

\subsubsection{Mask-Guided Dense Correspondence Sampling}
\label{subsubsec:mask_guided_matching}

Fisheye distortion, specular paint, and weakly textured body panels make sparse keypoints unreliable. We apply RoMA~v2.0.1~\cite{edstedt2025roma} with its public \texttt{precise} configuration to each complete, uncropped image pair $(I_A,I_B)$. The images remain in their calibrated camera projection and are not multiplied by object masks before inference; RoMA applies its standard internal resizing. This preserves global context for resolving repeated automotive structure while restricting only the correspondences exported to SfM.

RoMA predicts dense bidirectional warps, overlap probabilities, and directional $2\times2$ precision matrices. We use its released overlap-weighted, density-balanced sampler with a fixed proposal budget $N_s$ and a pair-deterministic seed. The returned overlap probability $o_i$ is used for admission and ranking; the precision matrices are not collapsed into an ad hoc scalar confidence. Privacy and non-rigid regions are excluded through the endpoint-admission mask
\begin{equation}
\overline M_k^{\mathrm{SfM}}
=
\operatorname{erode}
\left(
M_k^{\mathrm{rigid}}\wedge\neg P_k
\right),
\label{eq:sfm_endpoint_mask}
\end{equation}
where $P_k$ is the privacy mask. Thus, RoMA observes the complete RGB image, but both endpoints of every exported match must lie on the privacy-safe, wheel-excluded rigid body.

We reject correspondences between visually similar but geometrically different vehicle regions, $\psi_k(\mathbf x)=(\ell_k,v_k)\in[0,1]^2$ maps a pixel inside the oriented vehicle box to normalized rear-to-front and vertical coordinates. Cross-view matches must agree in both coordinates and collectively cover a fixed canonical vehicle grid. A capture-geometry-defined subset $\mathcal E_{\mathrm{bridge}}\subset\mathcal E$ contains the front, rear, and top connections required to join otherwise weakly overlapping viewpoint groups. These edges additionally require agreement with a separate swapped-order RoMA inference. The bridge labels and all operating points are fixed before matching and do not depend on the resulting reconstruction.

\begin{algorithm}[t]
\caption{Mask-Guided RoMA~v2 Matching}
\label{alg:roma_export}
\small 
\algsetup{indent=0.75em}
\begin{algorithmic}[1]
\REQUIRE Pair graph $\mathcal E$, bridge set $\mathcal E_{\mathrm{bridge}}$; images $\{I_k\}$; endpoint masks $\{\overline M_k^{\mathrm{SfM}}\}$; camera models $\{\pi_k\}$; canonical maps $\{\psi_k\}$; frozen operating point $\Theta$
\STATE Initialize persistent per-image keypoint registries $\mathcal K_k\leftarrow\varnothing$
\FOR{each $(A,B)\in\mathcal E$}
    \STATE $\mathcal R_{AB}\leftarrow
    \operatorname{RoMA}_{\mathrm{precise}}(I_A,I_B)$
    \STATE $\mathcal C\leftarrow
    \operatorname{Sample}_{N_s,\eta(A,B)}(\mathcal R_{AB})$
    \STATE $\mathcal C\leftarrow
    \{c_i\in\mathcal C:
    o_i\geq\tau_o,\
    \overline M_A^{\mathrm{SfM}}(\mathbf x_i^A)=
    \overline M_B^{\mathrm{SfM}}(\mathbf x_i^B)=1\}$
    \IF{$A$ and $B$ are different physical viewpoints}
        \STATE $\mathcal C\leftarrow
        \{c_i\in\mathcal C:
        |\ell_i^A-\ell_i^B|\leq\tau_\ell,\
        |v_i^A-v_i^B|\leq\tau_v\}$
    \ENDIF
    \IF{$(A,B)\in\mathcal E_{\mathrm{bridge}}$}
        \STATE $\mathcal C_{BA}^{\mathrm{swap}}\leftarrow
        \operatorname{Sample}
        (\operatorname{RoMA}_{\mathrm{precise}}(I_B,I_A))$
        \STATE $\mathcal C\leftarrow
        \{c_i\in\mathcal C:
        d_{\mathrm{swap}}(c_i,\mathcal C_{BA}^{\mathrm{swap}})
        \leq\tau_{\mathrm{cyc}}\}$
    \ENDIF
    \IF{$A$ and $B$ are different physical viewpoints and $\operatorname{Occ}_{\mathcal G_{\mathrm{can}}}(\mathcal C)<\kappa_{AB}$}
        \STATE \textbf{continue}
    \ENDIF
    \STATE $\mathcal C\leftarrow
    \operatorname{Unique}_{q}(\mathcal C;o)$
    \STATE $\mathcal C\leftarrow
    \operatorname{GridTopK}^{A,B}_{\mathcal G_{\mathrm{img}},K}
    (\mathcal C;o)$
    \STATE $\mathcal C\leftarrow\operatorname{Top}_{N_{\max}}(\mathcal C;o)$
    \STATE $(\widetilde{\mathcal C},\mathcal K_A,\mathcal K_B)
    \leftarrow\operatorname{Register}_{r}
    (\mathcal C,\mathcal K_A,\mathcal K_B)$
    \STATE $\mathcal V_{AB}\leftarrow
    \operatorname{CalibratedRANSAC}
    (\widetilde{\mathcal C};\pi_A,\pi_B)$
    \IF{$\operatorname{Accept}_{\mathrm{geo}}(\mathcal V_{AB})$}
        \STATE Export the verified registry-index pairs in $\mathcal V_{AB}$
    \ENDIF
\ENDFOR
\end{algorithmic}
\end{algorithm}

Here, $\eta(A,B)$ is the deterministic pair ordinal in the canonically sorted pair list. For a forward candidate, $d_{\mathrm{swap}}$ is the minimum, over reverse proposals after swapping their endpoint order, of the larger normalized-image endpoint distance; both endpoints must therefore agree with the same reverse proposal. $\operatorname{Occ}_{\mathcal G_{\mathrm{can}}}$ counts occupied cells of the fixed canonical vehicle grid only for cross-view edges; $\kappa_{AB}$ distinguishes ordinary cross-view connections from critical bridges, and same-view temporal edges bypass this test. $\operatorname{Unique}_{q}$ retains the highest-overlap candidate when quantized endpoint pairs collide, while $\operatorname{GridTopK}^{A,B}$ limits each original-image grid cell in both images so that a single reflection or textured component cannot dominate.

$\operatorname{Register}_{r}$ is a persistent per-image spatial registry: nearby observations reuse a stable keypoint identifier and its running-mean subpixel location, allowing independently processed image pairs to form multi-view tracks. Geometric verification back-projects each retained pixel through its calibrated camera model,

\begin{equation}
\mathbf b_i^A=\pi_A^{-1}(\mathbf x_i^A),\quad
\mathbf b_i^B=\pi_B^{-1}(\mathbf x_i^B),\quad
(\mathbf b_i^B)^\top\mathbf E\,\mathbf b_i^A=0,
\label{eq:feature_bearing_verification}
\end{equation}

\noindent
and exports a pair only when the predeclared calibrated-RANSAC inlier conditions are satisfied. The configured image-space tolerance is not applied directly to distorted pixels: each calibrated camera converts it to normalized camera units, and COLMAP averages the two converted thresholds before essential-matrix RANSAC. All settings are frozen before test reconstruction; Supplement Sec.~S3 gives their executed values.

\newpage
\subsection{Object-Centric SfM}
\label{subsec:methodology_structure_from_motion}

Incremental SfM can drift or fragment across wide baselines when constrained only by a specular moving vehicle. We therefore use global SfM~\cite{pan2024global} on the geometrically verified rigid-body correspondences.

For synchronized fixed-rig capture, let $o$, $r$, and $c$ denote the object, rig, and camera frames. The object-to-camera pose is
\begin{equation}
\mathbf T_{c\leftarrow o}(t)=
\underbrace{
\exp\!\left(\widehat{\boldsymbol\xi}_c\right)
\bar{\mathbf T}_{c\leftarrow r}
}_{\mathbf T_{c\leftarrow r}}
\mathbf T_{r\leftarrow o}(t),
\label{eq:rig_factorization}
\end{equation}
where $\mathbf T_{r\leftarrow o}(t)$ is the shared vehicle trajectory, $\bar{\mathbf T}_{c\leftarrow r}$ is the nominal CAD transform, and $\boldsymbol\xi_c\in\mathbb R^6$ is a bounded, time-invariant installation correction. Independent handheld passes instead receive unconstrained object-to-camera poses $\mathbf T_{c_i\leftarrow o}$. Let
$\mathcal Q=\{\mathbf T_{r\leftarrow o}(t)\}_t$ for fixed-rig capture or
\mbox{$\mathcal Q=\{\mathbf T_{c_i\leftarrow o}\}_i$} for handheld capture, and let
$\boldsymbol\Theta_i(\mathcal Q,\boldsymbol\xi)$ denote the resulting pose for image $i$.

Given verified observations $\mathcal O$, global mapping triangulates tracks and solves
\begin{align}
\min_{\{\mathbf X_j\},\mathcal Q,\{\boldsymbol\xi_c\}}
&\sum_{(i,j)\in\mathcal O}
\rho\!\left(
\left\|
\pi_{c(i)}\!\left(
\boldsymbol\Theta_i(\mathcal Q,\boldsymbol\xi)\mathbf X_j
\right)-\mathbf x_{ij}
\right\|_2^2
\right)
\notag\\[-2pt]
&+\mathbf 1_{\mathrm{rig}}\lambda_{\mathrm{rig}}
\sum_c
\boldsymbol\xi_c^\top
\boldsymbol\Sigma_c^{-1}
\boldsymbol\xi_c ,
\label{eq:object_centric_sfm}
\end{align}

\noindent
where $\mathbf X_j$ is an object-frame point, $\rho$ is a robust loss, $\pi_{c(i)}$ is the calibrated projection, and $\boldsymbol\Sigma_c$ bounds rig-pose refinement.


CAD translations establish fixed-rig metric scale; fixing $\mathbf T_{r\leftarrow o}(t_{\mathrm{ref}})$ anchors the $\mathrm{SE}(3)$ gauge. Handheld reconstruction retains a similarity gauge. LiDAR is evaluation-only. Rigid-mask admission suppresses background and wheel tracks. A predeclared audit checks registration, calibrated reprojection error, physical-viewpoint track span, within-pass continuity, and vehicle-shell/frusta coherence, without an absolute handheld cross-pass constraint. Accepted geometry is frozen across renderers. Supplement Secs.~S1.1 and~S3.3 detail the transform convention and audit, respectively.


\subsection{Rendering and Surface Evaluation}
\label{subsec:methodology_gaussian_splatting}

The recovered object-centric sparse cloud and frozen cameras initialize every renderer. Standard 3DGS~\cite{kerbl20233d} assumes pinhole projection, requiring pre-undistortion that resamples peripheral wide-FOV evidence. For native fisheye rendering, 3DGUT~\cite{wu20253dgut} instead propagates each Gaussian $\mathcal N(\boldsymbol\mu,\boldsymbol\Sigma)$ through the calibrated non-linear camera using Unscented Transform sigma points:
\begin{equation}
\mathbf X_0=\boldsymbol\mu,\quad
\mathbf X_k^{\pm}=\boldsymbol\mu\pm
\left[\sqrt{(d+\lambda_{\rm UT})\boldsymbol\Sigma}\right]_k,
\quad k=1,\ldots,d.
\label{eq:ut_sigma_points}
\end{equation}
Each $\mathbf X_s$ is projected as
$\mathbf u_s=\pi_c(\mathbf K_c,\boldsymbol\theta_c,
\mathbf T_{c\leftarrow o}(t),\mathbf X_s)$, and the weighted projections define the image-plane mean and covariance. 3DGUT+MCMC additionally uses stochastic densification for weakly textured and reflective regions~\cite{kheradmand20243d}.

Clean rendering preserves the source RGB and supplies object support separately. Let $M_0=M^{\mathrm{rigid}}$ for the common benchmark and $M_0=M^{\mathrm{render}}$ for the complete-mask ablation. Privacy is excluded through the valid domain $V=\neg P$, yielding
\begin{equation}
\begin{aligned}
F&=\operatorname{erode}(M_0)\cap V,\\
U&=\bigl(\operatorname{dilate}(M_0)\setminus
          \operatorname{erode}(M_0)\bigr)\cap V,\\
B&=\neg\operatorname{dilate}(M_0)\cap V .
\end{aligned}
\label{eq:privacy_valid_trimap}
\end{equation}
Every image-domain loss is restricted to $V$ and augments the renderer's native objective with foreground-normalized photometric supervision, foreground opacity or occupancy, confident-exterior suppression, and soft boundary consistency:
\begin{equation}
\mathcal L=
\mathcal L_{\mathrm{native}}
+\lambda_{\mathrm{rgb}}\mathcal L_F^{\mathrm{rgb}}
+\lambda_{\mathrm{fg}}\mathcal L_F^{\alpha}
+\lambda_{\mathrm{ext}}\mathcal L_B^{\alpha}
+\lambda_{\mathrm{bd}}\mathcal L_U^{\mathrm{bd}}.
\label{eq:clean_rendering_objective}
\end{equation}
Privacy pixels contribute neither training targets nor metric terms; Supplement Sec.~S4 specifies the validity-aware PSNR, SSIM, and LPIPS evaluation. For SSIM and LPIPS, prediction and reference are neutralized identically outside the privacy-valid object domain. Only windows or feature locations whose complete receptive fields lie inside that domain contribute to the reported score. RGB values are normalized to $[0,1]$. PSNR normalizes RGB squared error by $3|D|$, where $D$ is the privacy-excluded object domain; white exterior pixels do not enter this sum.

Gaussian methods additionally use multi-view support and capacity-aware densification admission to suppress poorly supported exterior primitives, while Instant-NGP applies analogous density, transmittance, and empty-space supervision. Native color and alpha remain separate, and white-background outputs are formed only at rendering time as
$\widehat I_{\mathrm{white}}
=\widehat\alpha\widehat I+(1-\widehat\alpha)\mathbf 1$;
alpha is never inferred from RGB or replaced by a binary mask. For premultiplied color $C=\widehat\alpha\widehat I$, the equivalent white composite is $C+(1-\widehat\alpha)\mathbf1$; multiplying $C$ by alpha again would apply opacity twice.

All methods share frozen poses, masks, training images, staggered core and buffer exclusions, scored frames, and output resolution while retaining their published representations, native objectives, and validated schedules. Fixed-rig 3DGUT and Fisheye-GS use calibrated fisheye projection; handheld methods use the common frozen normalized/undistorted inputs.

Reference LiDAR is used only for surface evaluation. We independently canonicalize it, estimate an upright proper-axis $\mathrm{Sim}(3)$, and refine the alignment using robust multiscale ICP constrained to preserve semantic up. For aligned model points $P$ and reference points $Q$, symmetric Chamfer-L1 is

\begin{equation}
d_{\mathrm{C,L1}}=
\tfrac12\biggl(
\tfrac1{|P|}\sum_{\mathclap{p\in P}}\min_{q\in Q}\lVert p-q\rVert_2
+\tfrac1{|Q|}\sum_{\mathclap{q\in Q}}\min_{p\in P}\lVert q-p\rVert_2
\biggr).
\label{eq:surface_metrics}
\end{equation}

\newpage\clearpage
\section{Experiments}
\label{sec:experiments}
\subsection{Data, Calibration, and Protocol}
\label{subsec:experiments_data_protocol}
\begingroup
\makeatletter
\renewcommand\paragraph{\@startsection{paragraph}{4}{\z@}{1.25ex \@plus .2ex \@minus .1ex}{-1em}{\normalfont\normalsize\bfseries}}
\makeatother
\paragraph{Fixed-rig corpus.}
The fixed-rig corpus contains 25 vehicles captured at 10 dealership installations across the United States. Every installation uses the same camera model and nominal 14-camera CAD layout; unit-to-unit manufacturing variation, mounting tolerances, illumination, and background clutter provide the deployment variation. A single reference ChArUco calibration and CAD-derived relative pose configuration are transferred across sites without per-site recalibration. The fisheye calibration has $0.42$-pixel RMS reprojection error at 4K resolution. We apply the configured pipeline and matched baseline matrix independently to every vehicle and report the aggregate comparisons in Tables~\ref{tab:sfm_comparison}--\ref{tab:component_ablation}.

\paragraph{Handheld corpus and release.}
The six-vehicle handheld corpus is publicly available on Zenodo\footnote{Dataset and updates: \href{https://doi.org/10.5281/zenodo.23620751}{doi:10.5281/zenodo.23620751}.}; each vehicle was recorded in 14 independent drive-through passes. The phone's central-radial intrinsics are calibrated by the same ChArUco procedure with $0.47$-pixel RMS reprojection error. Release videos use opaque tracked bounding-box masks for license plates, occupants, background license plates, and people. Audio streams and identifying metadata are removed, and every sequence passes a manual frame audit. 

\paragraph{Evaluation protocol.}
Each recording uses one deterministic staggered per-stream temporal-block split. A distinct non-overlapping interval of normalized visible-vehicle progress is assigned to each of the 14 fixed-rig camera streams or independent handheld passes. Within each interval, an interior core is scored and the immediately adjacent same-stream observations form an unscored exclusion buffer. The union of all core and buffer blocks is excluded from a single training run. For synchronized fixed-rig data, observations from the other cameras at the corresponding vehicle progress remain eligible for training. For handheld data, other independent passes at compatible normalized progress remain eligible, without assuming cross-pass temporal synchronization. Matching and SfM use the selected recording; renderer training excludes the core and buffer blocks. All renderers share the frozen split and SfM scaffold for appearance interpolation along the capture manifold.
\par\endgroup

\begin{figure}[t]
    \centering
    \includegraphics[width=\linewidth]{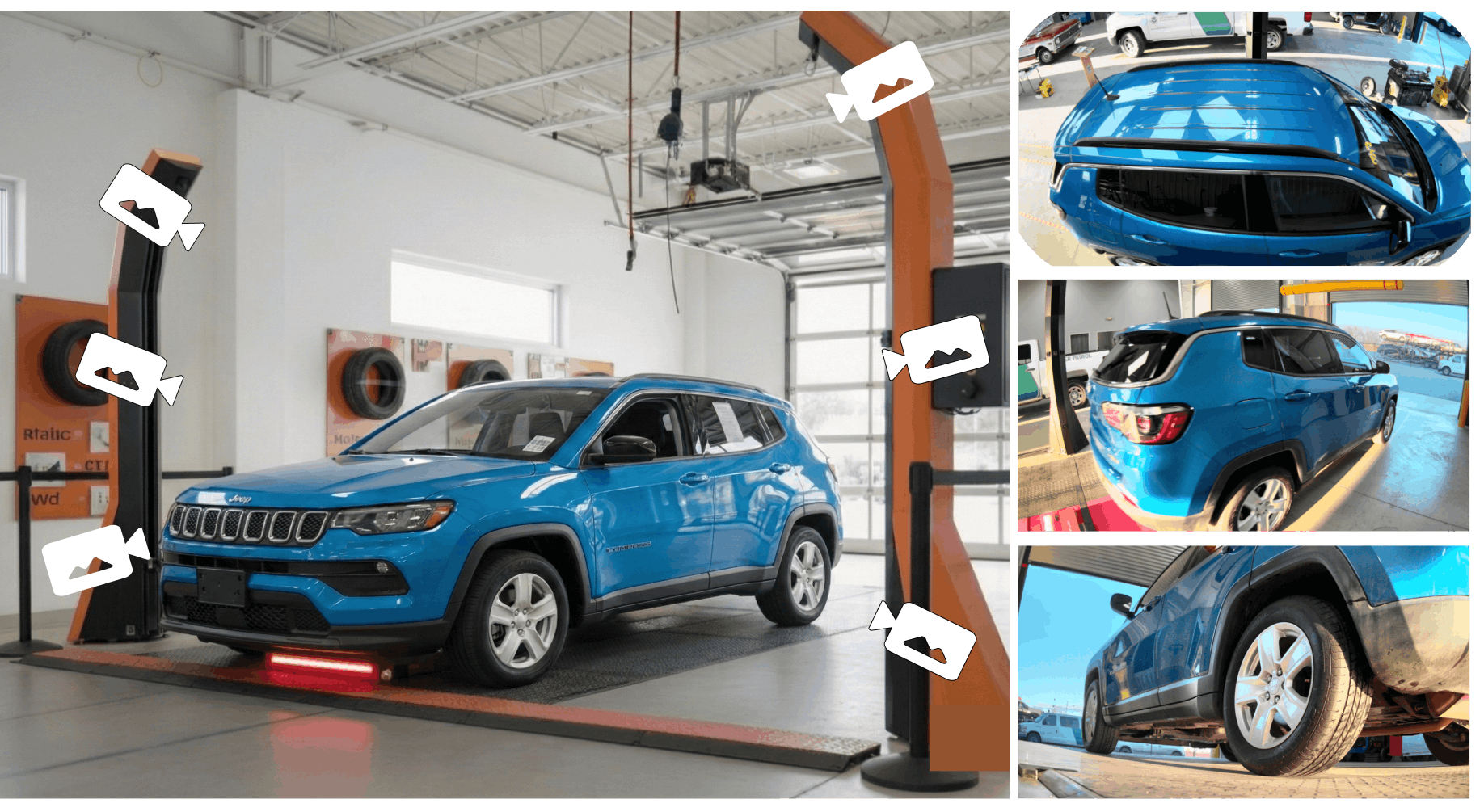}
    \caption{Fixed-rig data acquisition. Fourteen synchronized wide-angle cameras on two pillars observe a vehicle traversing a cluttered capture volume. The three vertical tiers and front-, side-, and rear-facing orientations provide complementary exterior coverage in one pass.}
    \label{fig:data_collection}
\end{figure}

\subsection{Rigid Vehicle Isolation and Tracking}
\label{subsec:experiments_rigid_vehicle_isolation_and_tracking}
\begin{figure}[t]
    \centering
    \includegraphics[width=\linewidth]{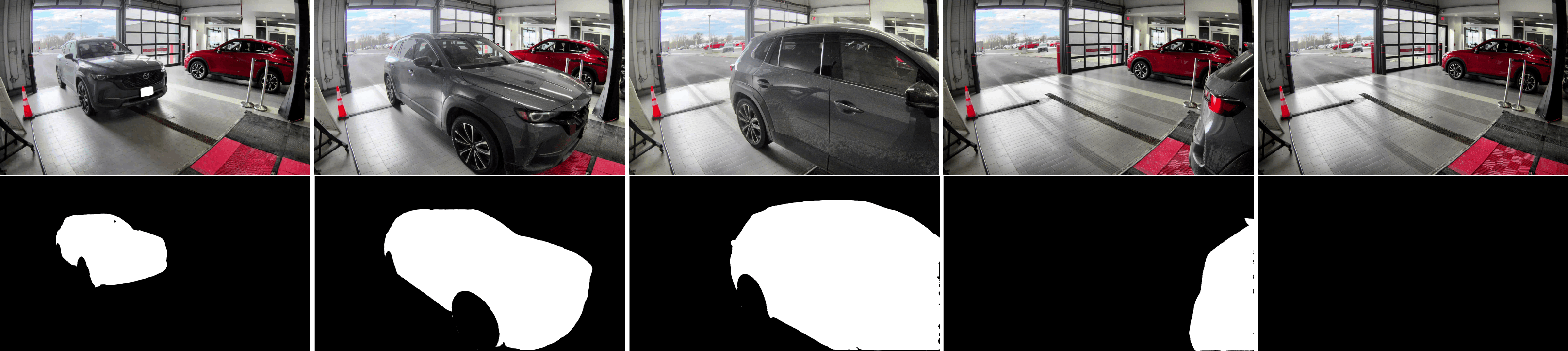}
    \caption{Rigid vehicle isolation and tracking. Top: representative raw frames. Bottom: geometry-valid masks produced by motion-gated tracking. The masks retain the rigid vehicle body while suppressing clutter and rotating wheels for matching and SfM.}
    \label{fig:rigid_vehicle_isolation_and_tracking}
\end{figure}

Figure~\ref{fig:rigid_vehicle_isolation_and_tracking} shows the temporal progression of rigid-body masks. Motion-gated SAM~3 identifies the moving target while suppressing dealership clutter and parked distractors. Before entry and after exit, anchor-locked tracking emits blank masks, preventing background correspondences from entering SfM.

The tracked silhouette is divided into rigid-body and complete appearance masks. Removing wheel regions before matching and SfM suppresses non-rigid wheel motion while retaining the approximately rigid body needed for reconstruction. Complete masks retain wheel appearance for the matched rendering ablation in Sec.~\ref{subsec:experiments_ablation}.

\subsection{Feature Matching}
\label{subsec:experiments_feature_matching}
\begin{figure}[t]
    \centering
    \includegraphics[width=\linewidth]{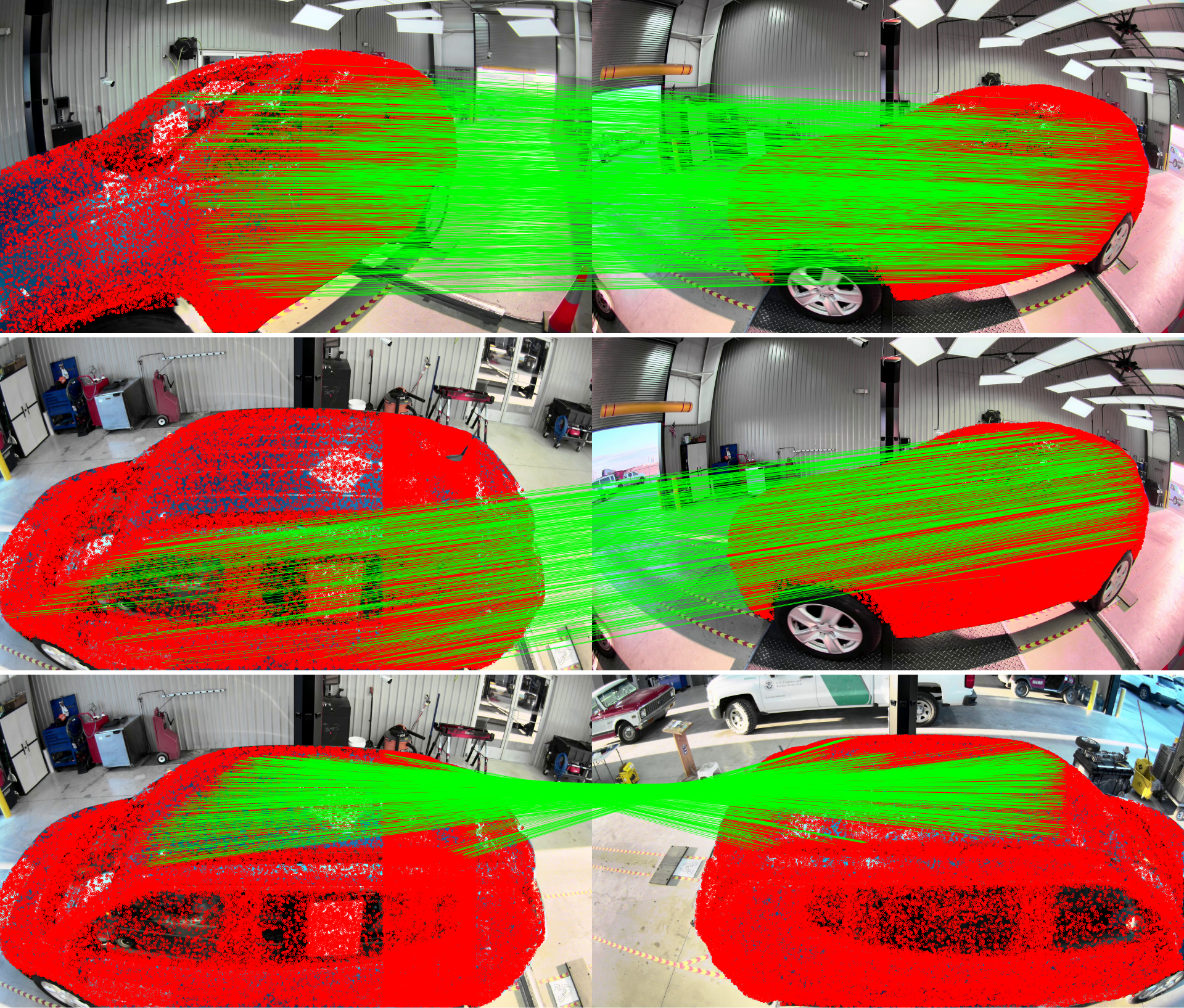}
    \caption{Mask-guided RoMA~v2 correspondences on raw distorted images. Dense matches retain full-image context while endpoint admission restricts SfM evidence to the rigid vehicle body across wide viewpoint changes.}
    \label{fig:roma_v2_matches}
\end{figure}

Figure~\ref{fig:roma_v2_matches} visualizes RoMA~v2 correspondences after rigid-body endpoint admission and confidence-aware geometric filtering. RoMA~v2 operates on the complete calibrated image pair, preserving global visual context across wide viewpoint changes. Accepted correspondences must place both endpoints within the rigid, wheel-excluded vehicle support, preventing stationary background structure and independently rotating wheel regions from entering SfM without hard-cropping the matcher input. Calibrated geometric verification subsequently rejects inconsistent correspondences. All matchers use the same images, masks, bounded pair graph, intrinsics, verification policy, and mapper; their resulting reconstruction outcomes are compared in Sec.~\ref{subsec:experiments_structure_from_motion}. The frozen exporter uses 5,000 proposals per pair, a 0.05 overlap gate, a $96{\times}96$-pixel balancing grid, and a 4-pixel keypoint-registry radius (Supplement Sec.~S3). Critical bridges require swapped-order agreement within 0.03 normalized image units; both endpoints must agree with the same reverse proposal. Persistent per-image keypoint identifiers link verified correspondences from independently processed pairs into multi-view tracks.

\newpage
\subsection{Object-Centric Structure-from-Motion}
\label{subsec:experiments_structure_from_motion}
We distinguish image registration from valid object-centric reconstruction.
For recording $s$, let $g_s=1$ only if the output forms one connected component
spanning all \mbox{14 physical cameras} or handheld passes, has positive non-degenerate vehicle support,
satisfies the frozen distortion-aware reprojection threshold, and passes the
frozen checks excluding background-locked, duplicated, or folded geometry.
We report the image-weighted valid registration coverage
$R=\sum_s g_s|\mathcal{V}^{(s)}_{\rm reg}|/
\sum_s|\mathcal{V}^{(s)}|$, together with mean reprojection error $E$,
mean track length $\bar L$, and sparse points $N$ over valid reconstructions.
Numerically completed models that fail these criteria contribute zero to $R$.
Table~\ref{tab:sfm_comparison} compares full 14-view reconstruction in both
capture regimes.


\begin{table}[t]
\centering
\footnotesize
\caption{Failure-aware full 14-view SfM comparison over \mbox{25 fixed-rig} and six handheld vehicles.
$R$: image-weighted valid registration coverage;
$E$: reprojection error (px);
$\bar L$: mean track length;
$N$: mean sparse points (millions).
$E$, $\bar L$, and $N$ use valid reconstructions only;
-- denotes that none met the validity criteria.}
\label{tab:sfm_comparison}
\setlength{\tabcolsep}{1.8pt}
\renewcommand{\arraystretch}{1.05}
\newcommand{\missingval}{\multicolumn{1}{c}{--}}
\begin{tabular*}{\columnwidth}{@{\extracolsep{\fill}}lrrrrrrrr@{}}
\toprule
& \multicolumn{4}{c}{Fixed rig}
& \multicolumn{4}{c}{Handheld} \\
\cmidrule(lr){2-5}\cmidrule(lr){6-9}
Method
& $R\uparrow$ & $E\downarrow$ & $\bar L\uparrow$ & $N\uparrow$
& $R\uparrow$ & $E\downarrow$ & $\bar L\uparrow$ & $N\uparrow$ \\
\midrule
SIFT
& 0.00 & \missingval & \missingval & \missingval
& 0.00 & \missingval & \missingval & \missingval \\
DISK+LG
& 0.00 & \missingval & \missingval & \missingval
& 0.00 & \missingval & \missingval & \missingval \\
LoFTR
& 0.00 & \missingval & \missingval & \missingval
& 0.63 & 1.11 & \textbf{3.77} & 0.38 \\
RoMA v2
& \textbf{1.00} & \textbf{0.61} & \textbf{4.01} & \textbf{0.96}
& \textbf{1.00} & \textbf{1.04} & 3.41 & \textbf{1.15} \\
\bottomrule
\end{tabular*}
\end{table}

\begin{figure}[t]
    \centering
    \includegraphics[width=\linewidth]{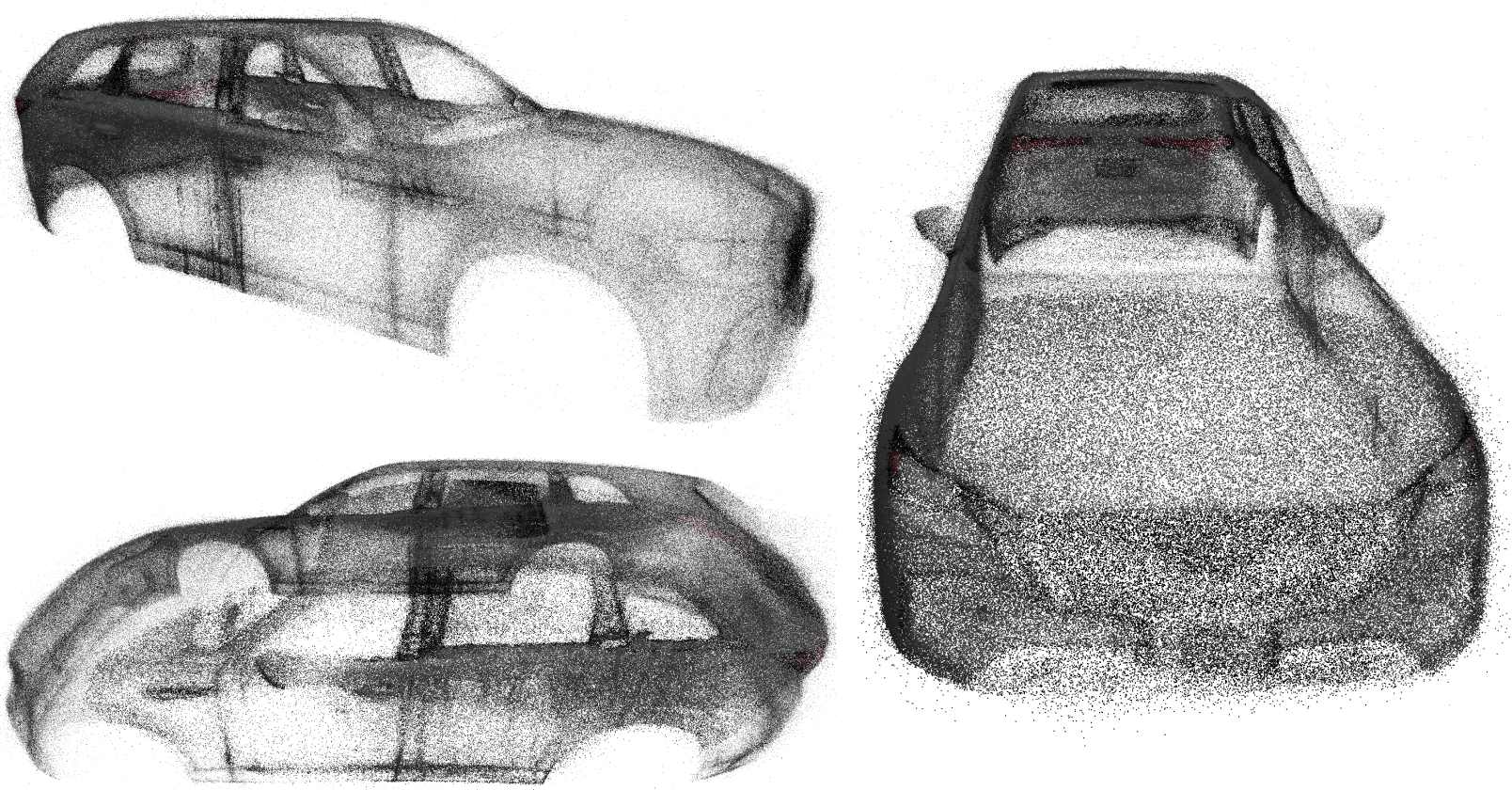}
    \caption{Representative successful object-centric sparse reconstruction from multiple viewpoints. The coherent vehicle shell and camera poses provide the common geometric scaffold for downstream rendering.}
    \label{fig:sparse_point_clouds_ours}
\end{figure}

RoMA~v2 reconstructs every recording and yields the lowest reprojection error in both regimes. SIFT and DISK+LG register images in some recordings, but their outputs fail the object-centric validity criteria. LoFTR succeeds on four of six handheld recordings but fails on the fixed rigs. Figure~\ref{fig:sparse_point_clouds_ours} shows a representative coherent vehicle shell.

\subsection{Rendering and Surface Evaluation}
\label{subsec:experiments_rendering}

\begin{figure*}[t]
  \centering
  \includegraphics[width=\textwidth]{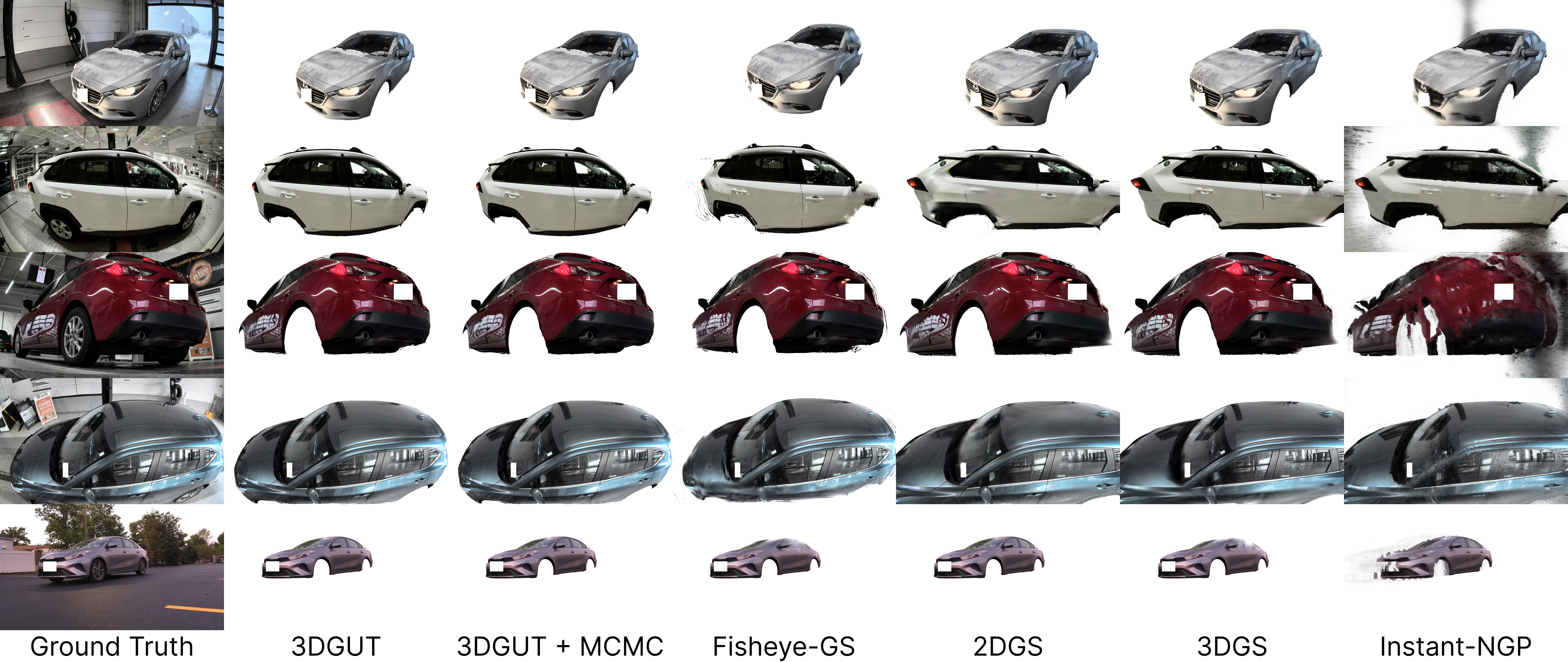}
    \caption{Qualitative comparison of novel-view vehicle rendering across diverse scenes. The leftmost column shows ground truth images; remaining columns show vehicle-centric renders from the compared methods.}
  \label{fig:ground_truth_vs_gaussian_splat}
\end{figure*}

Table~\ref{tab:rendering_comparison} uses the shared staggered split (Sec.~\ref{subsec:experiments_data_protocol}). Projection domains follow Sec.~\ref{subsec:methodology_gaussian_splatting}: fixed-rig 3DGUT and Fisheye-GS use calibrated fisheye projection, while handheld inputs are normalized/undistorted. Methods share frozen SfM, scored images, rigid/privacy support, trimap supervision, white-composite evaluation, and output resolution. Complete wheel-retaining support is isolated in the matched 3DGUT+MCMC ablation. Table~\ref{tab:rendering_comparison} reports held-out PSNR, SSIM, and LPIPS.

\begin{table}[t]
\centering
\scriptsize
\renewcommand{\arraystretch}{1.10}
\caption{Held-out rendering comparison on privacy-excluded rigid-body pixels using the shared staggered per-stream split. Metrics are averaged separately over the fixed-rig and handheld evaluations.}
\label{tab:rendering_comparison}
\resizebox{\columnwidth}{!}{%
\begin{tabular}{@{}lrrrrrr@{}}
\toprule
& \multicolumn{3}{c}{Fixed rig} & \multicolumn{3}{c}{Handheld} \\
\cmidrule(lr){2-4}\cmidrule(l){5-7}
Method & PSNR$\uparrow$ & SSIM$\uparrow$ & LPIPS$\downarrow$ &
PSNR$\uparrow$ & SSIM$\uparrow$ & LPIPS$\downarrow$ \\
\midrule
3DGS & 22.64 & 0.88 & 0.25 & 22.69 & 0.89 & 0.21 \\
3DGUT & 29.24 & 0.87 & 0.21 & 32.33 & 0.93 & 0.18 \\
3DGUT+MCMC & \textbf{30.42} & \textbf{0.90} & \textbf{0.20} & \textbf{32.71} & \textbf{0.94} & \textbf{0.17}\\
2DGS & 24.89 & 0.87 & 0.26 & 27.66 & 0.93 & 0.19 \\
Fisheye-GS & 20.50 & 0.86 & 0.28 & 22.33 & 0.88 & 0.23 \\
Instant-NGP & 20.53 & 0.76 & 0.43 & 22.40 & 0.79 & 0.35 \\
\bottomrule
\end{tabular}}
\end{table}


Reference LiDAR is available only for the handheld corpus and is never used for reconstruction or model selection. Table~\ref{tab:surface_comparison} reports full-surface and frozen body-only metrics after independent upright Sim(3)+ICP alignment. Instant-NGP uses one fixed density-to-surface extraction. 

\begin{table}[t]
\centering
\scriptsize
\renewcommand{\arraystretch}{1.10}
\caption{Full 14-view handheld surface comparison. Chamfer-L1 is in meters; F1 uses a 5-cm threshold.}
\label{tab:surface_comparison}
\resizebox{\columnwidth}{!}{%
\begin{tabular}{@{}lrrrr@{}}
\toprule
Method & Full C-L1$\downarrow$ & Body C-L1$\downarrow$ & Full F1$\uparrow$ & Body F1$\uparrow$ \\
\midrule
3DGS & 0.030 & 0.028 & 0.545 & 0.568 \\
3DGUT & \textbf{0.024} & 0.021 & 0.699 & 0.680 \\
3DGUT+MCMC & 0.026 & \textbf{0.013} & \textbf{0.703} & \textbf{0.885} \\
2DGS & 0.029 & 0.027 & 0.516 & 0.560 \\
Fisheye-GS & 0.027 & 0.014 & 0.694 & 0.877 \\
Instant-NGP & 0.214 & 0.212 & 0.262 & 0.253 \\
\bottomrule
\end{tabular}}
\end{table}

Body-only scores exclude the lowest 20\% of robust reference height and reuse the full-surface alignment. The two domains summarize the same reconstruction at different spatial extents.


\newpage
\begin{table}[!t]
\centering
\scriptsize
\caption{Full 14-view component ablations. $R$ is image-weighted valid registration coverage. $E$, $\bar L$, and $N$ use valid reconstructions only; -- denotes that no reconstruction satisfied the validity criteria, and N/A denotes an inapplicable ablation. The handheld pipeline has no rig-pose prior. Rendering-mask variants are compared using held-out LPIPS.}
\label{tab:component_ablation}
\setlength{\tabcolsep}{1.5pt}
\renewcommand{\arraystretch}{1.00}
\resizebox{\columnwidth}{!}{%
\begin{tabular}{@{}lrrrrrrrr@{}}
\toprule
& \multicolumn{4}{c}{Fixed rig}
& \multicolumn{4}{c}{Handheld} \\
\cmidrule(lr){2-5}\cmidrule(l){6-9}
SfM variant
& $R\uparrow$ & $E\downarrow$ & $\bar L\uparrow$ & $N$ (M)$\uparrow$
& $R\uparrow$ & $E\downarrow$ & $\bar L\uparrow$ & $N$ (M)$\uparrow$ \\
\midrule
Full pipeline
& \textbf{1.00} & \textbf{0.61} & \textbf{4.01} & 0.96
& \textbf{1.00} & \textbf{1.04} & \textbf{3.41} & \textbf{1.15} \\
w/o motion gate
& 0.55 & 1.03 & 2.69 & \textbf{1.60}
& \textbf{1.00} & \textbf{1.04} & 3.22 & 1.12 \\
w/o rigid endpoint masks
& 0.00 & -- & -- & --
& 0.00 & -- & -- & -- \\
w/o rig priors
& 0.00 & -- & -- & --
& \multicolumn{4}{c}{N/A} \\
SIFT+COLMAP
& 0.00 & -- & -- & --
& 0.00 & -- & -- & -- \\
\midrule
Mask variant
& \multicolumn{4}{c}{LPIPS$\downarrow$}
& \multicolumn{4}{c}{LPIPS$\downarrow$} \\
\midrule
Rigid-only
& \multicolumn{4}{c}{\textbf{0.20}}
& \multicolumn{4}{c}{\textbf{0.17}} \\
Bifurcated
& \multicolumn{4}{c}{0.23}
& \multicolumn{4}{c}{0.21} \\
\bottomrule
\end{tabular}}
\end{table}

\subsection{Ablations, Robustness, and Failures}
\label{subsec:experiments_ablation}
The matched ablations isolate motion gating, rigid endpoint masks, calibrated rig priors, correspondence estimation, and rendering-mask support. Table~\ref{tab:component_ablation} reports geometry ablations in both capture regimes and the rendering-mask ablation under the same full 14-view protocol. The rendering comparison holds the 3DGUT+MCMC poses, split, optimization schedule, objective, and capacity fixed, changing only whether wheel-retaining appearance support is included.

Motion gating raises fixed-rig valid coverage and reduces reprojection error, while handheld coverage remains unchanged. Removing rigid endpoint masks or fixed-rig pose priors produces background-centered or geometrically degenerate outputs. Remaining failures arise from disconnected camera components, background lock, duplicated or folded geometry, specular and textureless regions, severe occlusion, and wheel blur
or omission.

\section{Limitations and Ethical Considerations}
The private fixed-rig corpus has no metric surface reference, so Chamfer claims are restricted to the handheld set. Six handheld vehicles remain a modest transfer sample, and independent passes differ from synchronized fixed-rig capture. LiDAR alignment carries scale, sampling, and surrounding-geometry uncertainty; the reported full/body Chamfer-L1 and F1@5~cm values must be interpreted with these limitations. Reflective and textureless panels can produce local voids, and wheel retention may blur articulated regions. Strong exterior suppression can also trade fine silhouette detail for reduced floaters; the qualitative renders should therefore be considered alongside the held-out rendering metrics. The handheld release provides manually reviewed derivatives with opaque tracked bounding-box masks for license plates and visible people, including occupants; audio and identifying metadata are removed. The private dealership corpus is not distributed.

\section{Conclusion}
RigidDriveGS combines target-aware masking, raw-distorted dense matching, rig-constrained object-centric SfM, and clean projection-compatible rendering with separate trimap supervision and native-alpha compositing for fixed multi-camera drive-through capture. The method supports synchronized fixed-rig deployment and rig-free phone transfer, uses a support-preserving staggered temporal-block holdout across all 14 streams, and includes direct surface evaluation. Across both capture regimes, RoMA~v2 provides the strongest SfM connectivity, while 3DGUT+MCMC provides the strongest overall rendering and surface accuracy. Every renderer comparison uses the same frozen poses.

\newpage
\clearpage
{\small
\bibliographystyle{ieeenat_fullname}
\bibliography{main}
}
\end{document}